\documentclass[accepted,specialissue]{melba}

\usepackage{mwe} 

\usepackage{adjustbox}
\usepackage{amsmath,amsfonts}
\usepackage{mwe} 

\usepackage{amsmath,amsfonts}
\usepackage{booktabs}   
\usepackage{multirow}   
\usepackage{makecell}   
\usepackage{amsmath}    

\usepackage{float}     
\usepackage{tabularx}  
\usepackage{booktabs}   
\usepackage{array}     
\usepackage{ragged2e}  
\newcolumntype{Y}{>{\RaggedRight\arraybackslash}X}
\usepackage{siunitx}
\usepackage{dcolumn} 
\usepackage{booktabs}
\usepackage{colortbl}
 \usepackage{amsmath}
\usepackage{diagbox} 
\usepackage{booktabs}          
\usepackage{multirow}   

\melbaid{2025:035}  
\doi{https://doi.org/10.59275/j.melba.2025-d1g3}
\melbaauthors{Ghazal Danaee, Marc Niethammer, Jarrett Rushmore and Sylvain Bouix}  
\email{sylvain.bouix@etsmtl.ca}
\volume{3}
\firstpageno{792}  
\melbayear{2025}  
\datesubmitted{2025-04-21}  
\datepublished{2025-12-30}  

\melbaspecialissue{Special issue on Fairness of AI in Medical Imaging (FAIMI)}
\melbaspecialissueeditors{Veronika Cheplygina, Aasa Feragen, Andrew King, Ben Glocker, Enzo Ferrante, Eike Petersen, Esther Puyol-Antón, Melanie Ganz-Benjaminsen}

\ShortHeadings{Demographic Bias in Brain MRI Segmentation}{Danaee, Niethammer, Rushmore, Bouix}

\title{Investigating Demographic Bias in Brain MRI Segmentation: A Comparative Study of Deep-Learning and Non-Deep-Learning Methods}

\author{
	\firstname Ghazal \surname Danaee\aff{1}\orcid{0009-0000-0961-834X},
    \firstname Marc \surname Niethammer\aff{2}\orcid{0009-0003-7340-9050},
    \firstname Jarrett \surname Rushmore\aff{3},
    \firstname Sylvain \surname Bouix\aff{1}\orcid{0000-0003-1326-6054}
}
\affiliations{
\num 1 \addr École de technologie supérieure, Montréal, QC, Canada \\  
\num 2 \addr University of California San Diego, San Diego, CA, USA \\ 
\num 3 \addr Boston University School of Medicine, Boston, MA, USA 
}

\abstract{
	Deep-learning-based segmentation algorithms have substantially advanced the field of medical image analysis, particularly in structural delineations in MRIs. However, an important consideration is the intrinsic bias in the data. Concerns about unfairness, such as performance disparities based on sensitive attributes like race and sex, are increasingly urgent. In this work, we evaluate the results of three different segmentation models (UNesT, nnU‐Net, and CoTr) and a traditional atlas-based method (ANTs), applied to segment the left and right nucleus accumbens (NAc) in MRI images. We utilize a dataset including four demographic subgroups: black female, black male, white female, and white male. We employ manually labeled gold-standard segmentations to train and test segmentation models. This study consists of two parts: the first assesses the segmentation performance of models, while the second measures the volumes they produce to evaluate the effects of race, sex, and their interaction. Fairness is quantitatively measured using a metric designed to quantify fairness in segmentation performance. Additionally, linear mixed models analyze the impact of demographic variables on segmentation accuracy and derived volumes. Training on the same race as the test subjects leads to significantly better segmentation accuracy for some models. ANTs and UNesT show notable improvements in segmentation accuracy when trained and tested on race-matched data, unlike nnU-Net, which demonstrates robust performance independent of demographic matching. Finally, we examine sex and race effects on the volume of the NAc using segmentations from the manual rater and from our biased models. Results reveal that the sex effects observed with manual segmentation can also be observed with biased models, whereas the race effects disappear in all but one model. Our findings underscore the importance of diverse and balanced datasets for equitable brain MRI segmentation and highlight the need for systematic bias analysis in developing medical imaging models.
	}

\keywords{Bias, Fairness, Deep learning, Multi-atlas label fusion segmentation, Brain, MRI}

\begin{document}

\twocolumn[\maketitle]

\section{Introduction}
\enluminure{R}{esearchers} have widely adopted deep-learning-based models as state-of-the-art approaches in medical image computing. However, these models may display biased predictions for individuals with different protected attributes, such as sex, age, and race~\citep{Xu2024}.

When a model performs worse for specific subgroups, the downstream clinical implications can be significant, potentially leading to misdiagnosis or underdiagnosis for patients within those groups. Examining and mitigating these biases is paramount to achieving equitable healthcare outcomes. For instance, \citet{Stanley2022} studied differences in the performances of models predicting the sex of patients using magnetic resonance imaging (MRI) and observed differences between white and black children. \citet{Frazier2008} studied how diagnosis and sex affect brain regions in early-onset bipolar disorder and schizophrenia. Using MRI, they found that both factors influence amygdalar and hippocampal volumes, with differences between men and women. The study underscores the critical importance of accounting for sex differences in brain studies related to mental health.

Several factors contribute to bias in medical image computing. One fact is the inherent anatomical differences between men and women. In a recent study, \citet{Dibaji2024} evaluated how these sex-based anatomical differences in brain MRI data influence the performance of sex classification models. They analyzed saliency maps of the models to determine the regions most influential in driving sex classification. Furthermore, \citet{Isamah2010} demonstrated the differences in volumes of brain structures between various racial groups.

\indent Previous research has predominantly concentrated on fairness in classification tasks~\citep{mehrabi2022surveybiasfairnessmachine}. By contrast, fairness in segmentation has received relatively little attention, despite the significant impact that segmentation bias can have on clinical decision making. The few studies conducted in the realm of segmentation have typically focused on evaluating only one type of deep learning model in their analyses. We addressed these gaps by thoroughly evaluating demographic bias in both deep-learning and non-deep-learning models for brain region segmentation. Specifically, we considered four different methods: three state-of-the-art deep-learning models with different types of architectures (UNesT ~\citep{yu2023UNesT}, nnU‐Net~\citep{Isensee2021}, and CoTr~\citep{xie2021CoTrefficientlybridgingcnn}) and a traditional atlas-based segmentation method (Multi-Atlas Segmentation with Joint Label Fusion~\citep{ants}). We evaluated their bias across four demographic subgroups (black female, black male, white female, and white male). Moreover, we used manually annotated gold-standard segmentations of two subcortical structures—namely, the left and right nucleus accumbens (NAc)—as the labels for the training dataset, thus ensuring a high-quality gold standard for our evaluation. We extended our analyses beyond segmentation accuracy by investigating whether volume differences between sex, race and their interaction, observed with manual segmentation, remain consistent when using the segmentation output from biased models.\\
While recent studies have provided comprehensive analyses of bias across multiple attributes and mitigation strategies~\citep{siddiqui2024fair}, our work offers a distinct contribution by comparing the performance of multiple deep-learning architectures against a traditional non-deep-learning method in the context of brain MRI segmentation.\\
Ultimately, our findings aim to contribute to developing more equitable and generalizable practices in automated brain-image segmentation, thereby fostering enhanced fairness within both clinical and research environments.

\section{Related Works}
Previous studies have investigated bias in segmentation tasks in medical image computing. For example, Puyol-Antón et al. observed statistically significant differences in the performance of models in cine cardiac MR segmentation between racial groups~\citep{puyolanton2021fairnesscardiacmrimage, Puyol-Antón2022}. The training set was sex-balanced but not race-balanced, causing race bias in performance results. Additionally, \citet{lee2025investigation} showed racial biases in cine cardiac magnetic resonance (CMR) imaging with reduced performance on black subjects. They aimed to identify the causes of racial bias. They discovered that racial biases stemmed from non-cardiac features in MR images (areas of the image that did not include the heart), and training the model on cropped images helped narrow the performance gap between black and white patients. 

In a recent study, \citet{ioannou2022studydemographicbiascnnbased} investigated the influence of sex and race on the performance of the FastSurferCNN model~\citep{Henschel2020} trained using silver standard labels derived from the Multi-Atlas Label Propagation with Expectation-Maximization-based refinement (MALPEM) algorithm~\citep{Ledig2015, Ledig2018}. The study focused on the segmentation of 78 structures in the brain and evaluated demographic biases within these regions; they found sex and race bias in some but not all structures. To assess sex bias, they trained five different models on training sets with varying ratios of white females and white males and subsequently tested these models on the test sets of white females and white males. In another experiment, they used the same trained models and tested them on black and white females to measure race biases. Their findings revealed that race bias is more significant than sex bias. In addition, they reported that specific brain regions showed a significant bias effect, indicating that the bias has a spatial component. They also observed sex biases. For example, when the model was trained on a sex-balanced dataset, its performance in segmenting three brain regions showed a statistically significant reduction in Dice coefficient for white females compared to white males. \\

\citet{alqarni2024investigation} investigated racial bias in deep learning-based prostate gland segmentation from MR images by training models with varying ratios of white and black subjects. Their findings revealed that models trained on imbalanced datasets exhibited significant racial bias, while employing a race-balanced training set resulted in the best segmentation performance across both groups. In a recent study,~\citet{siddiqui2024fair} investigated biases related to age, sex, and race in the segmentation of hip and knee X-ray images. Their work demonstrates the trade-off between fairness and accuracy by comparing several bias mitigation strategies applied to U-Net models with different CNN backbones (ResNet18 and EfficientNet-B0). A study evaluating skin color bias in skin lesion segmentation algorithms (CNNs) by \citet{bencevic2024skincolorbias} observed lower performance on darker skin tones. Although they used mitigation methods, none of them were effective at reducing the bias.
In this study, we supplement these initiatives by directly comparing multiple methods and using a manually curated gold-standard dataset, therefore providing a more comprehensive view of how different methods handle unbalanced training data.\\
Prior studies of bias in segmentation have often evaluated a single deep-learning architecture within a given application. Outside the brain, comparative work in cardiac MR has shown that model choice itself can affect measured sex bias and race bias~\citep{Lee2023Cardiac}. In brain MRI, however, we are not aware of prior studies that jointly compare deep learning and atlas-based segmentation with respect to bias.\\

Our study offers a more comprehensive evaluation by (i) comparing two distinct types of segmentation models—deep learning–based and traditional non–deep learning approaches, a comparison not previously worked on, (ii) employing manually curated gold-standard labels for the nucleus accumbens, and (iii) examining both segmentation performance fairness and its impact on volumetric analyses through linear mixed-effects models.
\section{Methods}
\subsection{Data}
We utilized data from the Human Connectome Project (HCP) Young Adult dataset. According to the ``WU–Minn HCP 1200 Subjects Data Release: Reference Manual'', each HCP participant is given a participant identification number for tracking and asked a number of demographic questions, including gender, age, twin status (including self-reported zygosity), race, ethnicity, educational level, household income, and relationship status{\citep{hcpmanual}}.\\
The T1-weighted MRIs have a resolution of \( 260 \times 311 \times 260 \) voxels with an isotropic voxel spacing of \( 0.7  \). The subjects' ages ranged from 22 to 35.  Reporting race and ethnicity in this study was mandated by the NIH, consistent with the Inclusion of Women, Minorities, and Children policy. Sex, race and ethnicity were self-reported by participants. For the training phase, we employed 30, 32, 33, and 31 images corresponding to black female, black male, white female, and white male subjects, respectively. In the testing phase, we utilized 19, 20, 19, and 20 images for black female, white male, white female, and black male subjects.

\begin{figure}[htbp]
    \centering
    \includegraphics[width=0.5\linewidth]{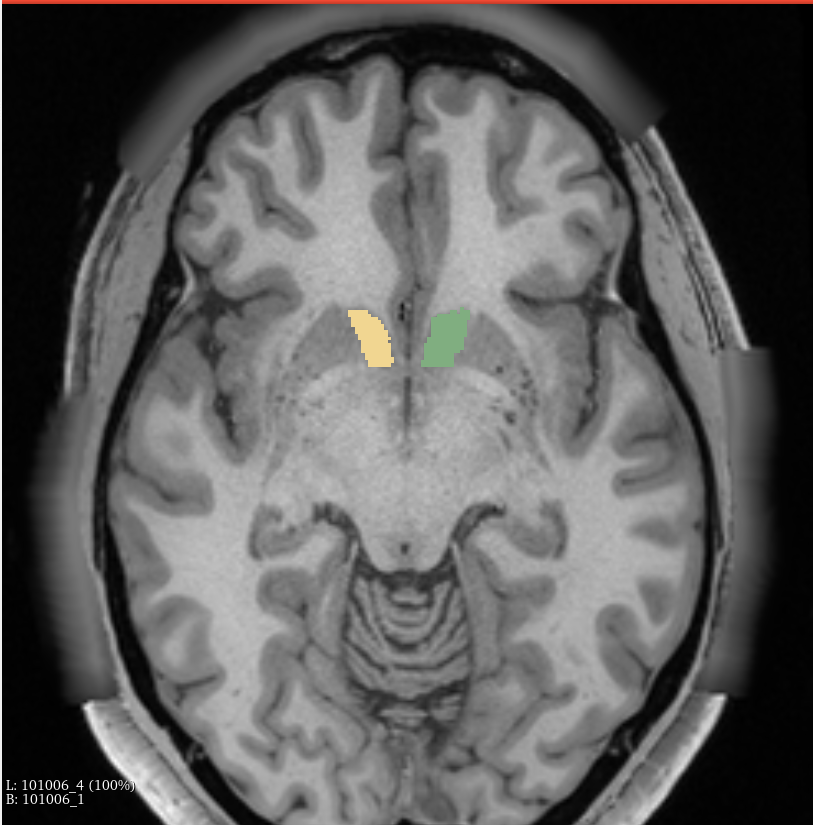}
    \caption{The green structure is Right NAc and the left structure is left NAc}
    \label{fig:NAc-photo}
\end{figure}

We utilized the manual segmentations of two subcortical structures, the right and left nucleus accumbens (NAc), provided by neuroanatomist Dr. Jarrett Rushmore. This structure was selected due to previously reported sex differences in microstructure \citep{wissman2011NucleusAccumbens}. Because NAc volume is widely employed as a volumetric biomarker, demographic bias in its automated segmentation could confound clinical inference and exacerbate health disparities. 
Figure~\ref{fig:NAc-photo} shows a case of manual annotation of the right and left NAcs.

\subsection{Biased training}
For each architecture—nnU-Net, UNesT, and CoTr—four separate models were trained, with each model using data from just one of the four demographic groups: 32 black male, 30 black female, 31 white male, or 33 white female subjects. Similarly, four biased datasets were used with ANTs, leading to 4 "models" each using atlases from just one demographic subgroup: 10 black male, 10 black female, 10 white male, or 10 white female subjects. This approach intentionally introduces bias to assess the impact of imbalanced training on segmentation fairness across demographics.
\subsection{Evaluation metrics}
We used two core metrics for evaluating raw segmentation performance.
First, the Dice similarity coefficient (DSC) is an overlap-based metric that ranges from 0 with no overlap to 1 with complete overlap. Let $X$ denote the ground truth segmentation and $Y$ be the predicted segmentation. The Dice coefficient is computed as:\\
\begin{equation}
\text{DSC}(X, Y) = \frac{2 |X \cap Y|}{|X| + |Y|}
\end{equation}
Second, the normalized surface Dice (NSD) defined by \cite{Nikolov2021} is a boundary-based metric that measures the Dice coefficient on boundary pixels with a margin. $\tau$ is the maximum tolerated distance from the boundary that defines the border region. For two shapes X and Y, $S_X$ and $S_Y$ denote their boundaries, and $\mathcal{B}_X^{(\tau)}$ and $\mathcal{B}_Y^{(\tau)}$ are the corresponding boundary regions~\citep{Maier-Hein2024}.
\begin{equation}
\text{NSD}(X,Y) = \frac{|S_X \cap \mathcal{B}_Y^{(\tau)}| + |S_Y \cap \mathcal{B}_X^{(\tau)}|}{|S_X| + |S_Y|}
\label{NSD}
\end{equation} 

\noindent Furthermore, to evaluate fairness in the models' results, we utilized the \textbf{Equity-Scaled Segmentation Performance (ESSP) metric}, originally proposed by \citet{tian2024fairseglargescalemedicalimage}. Computing ESSP requires two components:
\begin{enumerate}
    \item The overall segmentation accuracy (the mean accuracy of a model on all demographic groups)
    \item The deviation of a model's accuracy on each demographic group from the overall segmentation accuracy
\end{enumerate} 
Let $A$ denote the set of demographic groups under consideration (in our case, white male, white female, black male, and black female), $N_A$ denote the total size of all test sets across all subgroups in $A$, $\text{DSC}_a$ represent the average Dice coefficient for subgroup $a \in A$, and $n_a$ is the size of subgroup a. We compute the overall segmentation accuracy for each of the models by averaging the Dice coefficient over all samples in the test set:
\begin{equation}
\text{DSC}_{overall}=\frac{1}{N_A} \sum_{a \in A} \text{DSC}_a \times n_a.
\end{equation}

\noindent Next, we define $\Delta$ as the sum of absolute performance discrepancies across all groups:
\begin{equation}
\Delta = \sum_{a \in A} \Bigl| \text{DSC}_{overall} - \text{DSC}_a \Bigr|.
\end{equation}
Finally, we calculate the Equity-Scaled Segmentation Performance (ESSP) by penalizing the overall performance concerning $\Delta$:
\begin{equation}
\text{ESSP} = \frac{\text{DSC}_{overall}}{1 + \Delta}.
\end{equation}
In essence, ESSP acts as a substitute for the Dice coefficient, with a penalty for unfairness. 
Note that the same measure can be computed for NSD and we use the following notation $ESSP_{DSC}$ and $ESSP_{NSD}$ to respectively identify the Dice and NSD ESSP measures.

\subsection{Segmentation algorithms tested}
All deep-learning methods were used with their default configuration, as provided by their respective official code repositories. We chose the default configuration to mirror common practice, enable reproducible comparisons, and model real-world usage. There are therefore, some noticeable differences beyond network architectures, including the number of epochs and data augmentation. We summarize the training configurations of the deep learning models in Table~\ref{table:train-configs-portrait} and review each method in details below.

\subsubsection{UNesT}
We utilized UNesT~\citep{yu2023UNesT}, a segmentation model with a hierarchical transformer encoder that processes volumetric data by dividing the input into 3D patches and applying local self-attention at different scales. Unlike prior approaches relying on convolutional layers for feature extraction, UNesT leverages a transformer-based encoder to capture multi-scale features. It then uses a convolutional decoder to upsample these representations and produce the final segmentation.

\textbf{Implementation details:} We utilized the official implementation of the UNesT model and trained the UNesT-large version with approximately 280 million parameters from scratch on our dataset. We decided to use the default configuration of UNesT since these values worked best for our task of segmenting small subcortical structures. For example, UNesT could not segment well when trained with only 1000 epochs, which is the fixed value for the number of epochs in nnU-net. The code for UNesT is publicly accessible online\footnote{Official UNesT implementation: \url{https://github.com/MASILab/UNesT}}.\\
Data were registered to MNI space using affine transformations for the train and test phases. Then, after the test set was segmented, the results were registered back to the original space.
For each model trained on a specific demographic group (e.g., black females), we performed 5-fold cross-validation within that group’s training data, with each fold trained for 50,000 epochs, and the ensemble of the results of models were used to produce the predictions. We used the default parameters of the UNesT-large model, as provided by the official implementation. These included a learning rate of 0.00001, the Adam optimizer, and a momentum of 0.9. We utilized Dice-cross entropy as it showed better performance in segmenting small structures.

\subsubsection{nnU-Net}
nnU-Net~\citep{Isensee2021} is an adaptive model specifically designed for biomedical image segmentation. Its key advantage lies in its ability to apply to any dataset by systematizing the complex process of manual method configuration.

\textbf{Implementation details:} We trained the model from scratch on our training dataset, and evaluated its performance on test sets from different demographic groups. The official implementation of nnU-Net, used in our experiments, is available on their Github.\footnote{Official nnU-Net implementation: \url{https://github.com/MIC-DKFZ/nnUNet/tree/master/nnunetv2}}\\
We adhered to the default configuration for the nnU-Net model, as its defining characteristic is the automated optimization of the entire pipeline based on a new dataset's properties. Manually altering these systematically configured parameters would undermine the model's self-configuring design philosophy and negate its primary advantage.\\
nnU-Net adopts several fixed design choices, including the use of a combined cross-entropy and Dice loss function across all applications. In addition, it incorporates a set of rule-based and empirical design choices for the model's configuration.
Given a new training dataset, nnU-Net automatically creates up to three pipeline configurations (2D U-Net, 3D full-resolution U-Net, and the 3D U-Net cascade) , and trains each configuration in a 5-fold cross-validation run. After training, nnU-Net empirically selects the best single configuration or an ensemble of configurations~\citep{Isensee2021}. In our experiments, the 3D full-resolution configuration was consistently selected, and its predictions were used for reporting results.
nnU-Net’s rule-based and empirical strategies determined the other design choices.

\subsubsection{CoTr}
The last deep-learning-based model we used is CoTr~\citep{xie2021CoTrefficientlybridgingcnn}, which leverages the strengths of both transformers and convolutional neural networks for 3D medical image segmentation. In CoTr, a CNN is designed to extract feature representations, while the authors introduced the deformable Transformer (DeTrans) to model long-range dependencies within the extracted feature maps effectively.

\textbf{Implementation details:} We trained CoTr from scratch on the training set and evaluated it on test sets from various demographic groups.
We performed 5-fold cross-validation and used inference-time ensembling of the models. The loss function of the model is the sum of the Dice loss and cross-entropy loss. We used the official implementation of CoTr and its default configuration for training, which is available on GitHub.\footnote{Official CoTr implementation: \url{https://github.com/YtongXie/CoTr/tree/main}}

\begin{table*}[htbp]
\centering
\caption{Training configurations for deep learning models.}
\label{table:train-configs-portrait}
\renewcommand{\arraystretch}{1.4} 
\footnotesize 
\begin{tabularx}{\linewidth}{>{\raggedright\bfseries}p{2.5cm} Y Y Y}
\toprule
\textbf{Configuration} & \textbf{nnU-Net} & \textbf{CoTr} & \textbf{UNesT} \\
\midrule
\textbf{Loss function} 
    & Cross-entropy + Dice 
    & Cross-entropy + Dice 
    & Cross-entropy + Dice \\
\addlinespace
\textbf{Optimizer} 
    & SGD + Nesterov ($\mu=0.99$) 
    & SGD + Nesterov ($\mu=0.99$) 
    & Adam ($\beta_{1}=0.9$) \\
\addlinespace
\textbf{Learning rate (schedule)} 
    & 0.01 (poly: $(1 - \frac{\text{epoch}}{\text{epoch}_{\max}})^{0.9}$) 
    & 0.01 (poly: $(1 - \frac{\text{epoch}}{\text{epoch}_{\max}})^{0.9}$) 
    & $1\times 10^{-5}$ (warmup–cosine) \\
\addlinespace
\textbf{Epochs} 
    & 1000 
    & 1000 
    & 50,000 \\
\addlinespace
\textbf{5-fold CV} 
    & Yes 
    & Yes 
    & Yes \\
\addlinespace
\textbf{Preprocessing} 
    & Crop to non-zero region 
    & Same as nnU-Net 
    & Convert to MNI-305 space, reorient to RAS, resample, scale intensity, spatial pad \\
\addlinespace
\textbf{Data augmentation} 
    & Rotations, scaling, Gaussian noise/blur, brightness, contrast, low-res simulation, gamma correction, mirroring 
    & Same as nnU-Net 
    & Random patch sampling, random mirror flips, random multiplicative intensity scaling \\
\addlinespace
\textbf{Post-processing} 
    & Remove all but largest component 
    & Same as nnU-Net 
    & Convert back to original space \\
\bottomrule
\end{tabularx}
\end{table*}

\subsubsection{Multi-Atlas Segmentation with Joint Label Fusion (ANTs)}
Atlas-based segmentation relies on atlases—expert-labeled sample images— for guiding segmentation. Each atlas is registered to the target image in this approach, and the warped atlases are combined using label fusion techniques, such as weighted voting. Multi-Atlas Segmentation with Joint Label Fusion~\citep{ants} which incorporates dependencies between atlases, was one of the leading segmentation techniques before deep learning methods were developed. The method is quite flexible. Given a relatively small set of labeled data, one could perform segmentation with good accuracy. 

\textbf{Implementation details:} Four variants of ANTs Joint Label Fusion were used, each using 10 atlases exclusively from training sets of each one of the demographic subgroups: 10 black male, 10 black female, 10 white male, or 10 white female subjects to segment the test set. Since ANTs prediction entails cost-intensive atlas registration, we decided to choose 10 for the size, as in ablation studies with varying numbers of atlas subjects \((5, 10, 15, 20)\), we found that the segmentation accuracy derived by atlases curated with 10 cases differed from those curated with larger numbers of cases by less than 0.1 in Dice coefficient. We utilized the script provided by the Advanced Normalization Tools (ANTs) ecosystem, which is available on ANTs GitHub.\footnote{Official ANTs implementation: \url{https://github.com/ANTsX/ANTs/blob/master/Scripts/antsJointLabelFusion.sh}}
\subsection{Statistical analysis}
\subsubsection{Performance Bias}
In addition to evaluating DSC, NSD, ESSP and $\Delta$, we employed linear mixed models to assess bias in model performance. For each subject in the test sets of different demographic subgroups, we kept the performance scores from the four models within a single architectural design (e.g., the results of four UNesT models trained on black male, black female, white male, and white female). We then used the linear mixed effects model below on these performance scores (Dice coefficient):
\begin{multline}
    \text{DSC}= \beta_0 + \beta_1(\text{SameRace}) + \beta_2(\text{SameSex})\\
    +\beta_3(\text{SameRace} \times \text{SameSex} ) 
    + \epsilon
\end{multline}
where $\text{SameSex}$ is a binary variable which defines whether the test subject has the same sex as the training dataset, and $\text{SameRace}$ is a binary variable which defines whether the test subject has the same race as the training dataset. (e.g., coded as 1 for a match and 0 for a mismatch). The variables $\beta_0$, $\beta_1$, $\beta_2$, and $\beta_3$ are the parameters to be estimated, and $\epsilon$ is the error. This framework enabled us to quantify the contribution of each factor (as well as their interaction) to the observed Dice scores.
\subsubsection{Effect of bias on demographic analyses}
To compare the impact of a biased model on brain morphometry population analyses, we applied a linear mixed model to the volumes corresponding to the test sets of all demographic groups, produced by a single model from each architectural design and the demographic group utilized for training (e.g, UNesT model trained on black females). We then used the following linear mixed effects model on these volumes:
\begin{equation}
         \text{Volume}= \gamma_0 + \gamma_1(\text{Race}) + \gamma_2(\text{Sex}) +\gamma_3(\text{Race} \times \text{Sex} ) + \epsilon_2
\end{equation}
We can investigate how race, sex, and their interaction influenced the predicted volumes. $\text{Sex}$ and $\text{Race}$ are binary variables: $\text{Race}$ indicates whether the subject is black or white, and $\text{Sex}$ indicates whether the subject is female or male (e.g., White=0, Black=1; Male=0, Female=1).\\
To test whether age affected the volumes of manually annotated labels, we used the linear mixed-effects model below:
\begin{multline}
         \text{Volume}= \alpha_0 + \alpha_1(\text{Race})+ \alpha_2(\text{Sex}) +\alpha_3(\text{Sex $\times$ Race} ) \\
         + \alpha_4(\text{Age} ) +\epsilon_3
\label{eq:sex-race-age-LMM}
\end{multline}

\section{Results}
The following sections use model-subgroup notation. For instance, 'UNesTBF' represents the UNesT model trained with the black female subset.
\subsection{General statistics of the volumes}
The results of the linear mixed model showed that after adjusting for sex and race, the estimated effect of age on the volume of the nucleus accumbens (NAc) was not statistically significant. The results are provided in Table~\ref{table:sex-race-age-effect-manual-combined}. We therefore did not incorporate age as a variable of interest in any experiments. This is also supported by the fact that the HCP dataset has a rather narrow age range of 22 to 35 years. These results are also used in section \ref{sec:morphometricanalyses} to evaluate whether these relationships remain observable when tested on segmentations generated by automated methods.

\begin{table*}[htbp]
    \centering
    \caption{Results of the linear mixed model for evaluating sex, race, age, and their interaction (sex $\times$ race) effects on volumes by {\bf manual annotation} for right and left NAc. These results are also used to evaluate whether these relationships remain observable when tested on segmentations generated by automated methods.}
    \footnotesize
    \setlength{\tabcolsep}{4pt}
    \resizebox{\textwidth}{!}{%
    \begin{tabular}{ll ccc ccc ccc ccc}
        \toprule
        \multirow{2}{*}{\textbf{Structure}} & \multirow{2}{*}{\textbf{Method}} 
        & \multicolumn{3}{c|}{\textbf{Sex}} 
        & \multicolumn{3}{c|}{\textbf{Race}} 
        & \multicolumn{3}{c|}{\textbf{Sex $\times$ Race}} 
        & \multicolumn{3}{c}{\textbf{Age}} \\
        \cmidrule(lr){3-5} \cmidrule(lr){6-8} \cmidrule(lr){9-11} \cmidrule(lr){12-14}
        & & \boldmath{$\alpha_2$} & \textbf{Std Err} & \textbf{P-value} 
          & \boldmath{$\alpha_1$} & \textbf{Std Err} & \textbf{P-value} 
          & \boldmath{$\alpha_3$} & \textbf{Std Err} & \textbf{P-value}
          & \boldmath{$\alpha_4$} & \textbf{Std Err} & \textbf{P-value} \\
        \midrule
        \multirow{2}{*}{Right NAc} &
        Manual (whole dataset) & 195.38  & 69.39  & \textbf{0.005} 
                                & 231.38  & 69.76  & \textbf{0.001} 
                                & -62.279      & 97.133     & 0.521 
                                & -11.568 & 6.33   & 0.068 \\
        & Manual (Test set)      & 181.554 & 99.73  & 0.069 
                                 & 394.84  & 104.80 & \textbf{0.000} 
                                 & -98.103      & 143.006     & 0.493 
                                 & -6.422  & 5.960  & 0.281 \\
        \midrule
        \multirow{2}{*}{Left NAc} &
        Manual (whole dataset) & 222.831 & 67.218 & \textbf{0.001} 
                               & 257.216 & 67.573 & \textbf{0.000} 
                               & 5.810      & 94.088     & 0.951 
                               & -8.598  & 6.141  & 0.161 \\
        & Manual (Test set)      & 194.65  & 104.15 & 0.062 
                                 & 409.01  & 115.12 & \textbf{0.000} 
                                 & -94.511      & 162.562     & 0.561 
                                 & -9.916  & 6.733  & 0.141 \\
        \bottomrule
    \end{tabular}
    }
    \label{table:sex-race-age-effect-manual-combined}
\end{table*}

\begin{figure*}[htp]
    \centering
    \includegraphics[width=1.0\linewidth]{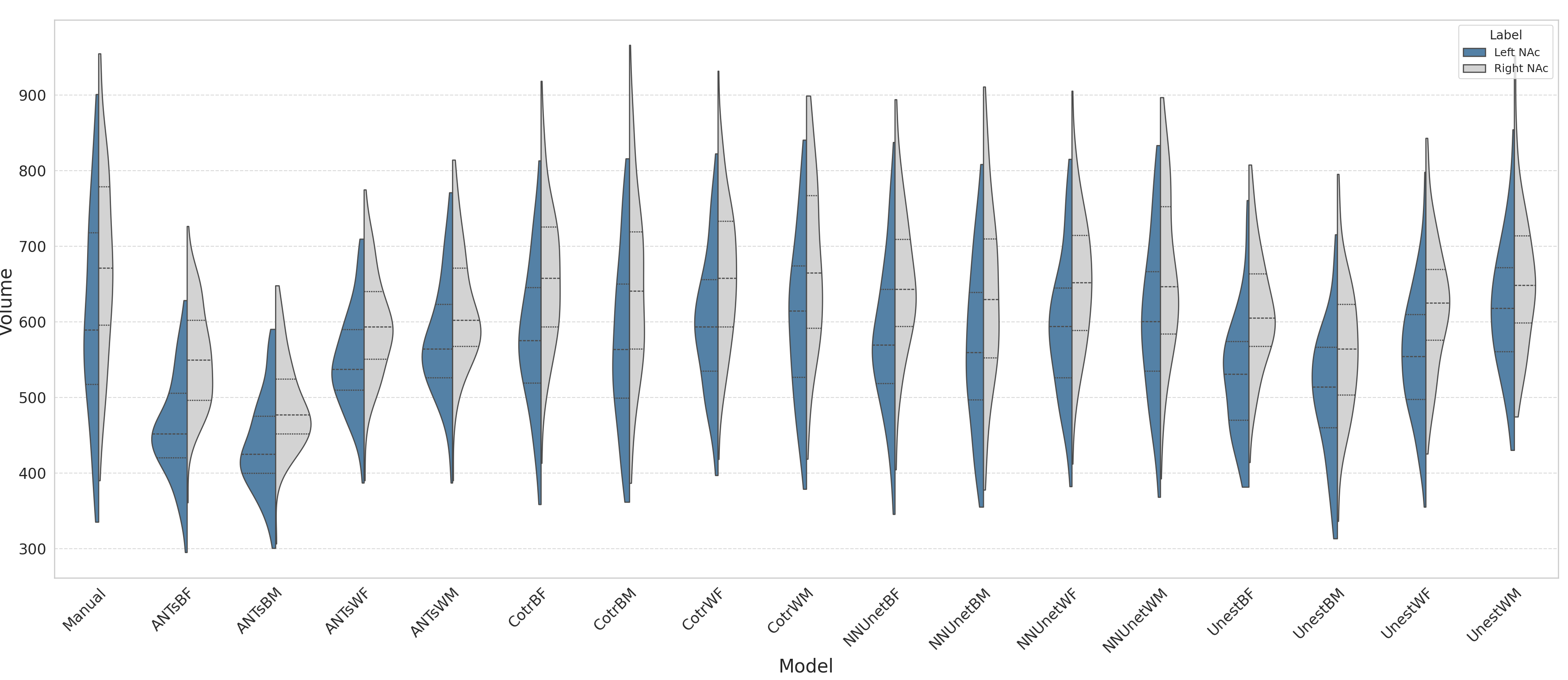}
    \caption{Left and right NAc volumes in \SI{}{\milli\meter\cubed} across manual and automated segmentations}
    \label{fig:volumes-splitviolin}
\end{figure*}

Table \ref{table:new-volumes-combined} displays volume statistics for all models. The ANTsBF and ANTsBM methods demonstrate greater under-segmentation than others. Notably, these are the only models with median volume differences below 20\% for the left NAc.
This is also observable in Figure \ref{fig:volumes-splitviolin} visualizing the volumes of the structures by manual annotations and models.

Most segmentation models exhibit smaller standard deviations compared to the manual approach. It could suggest that they are under-representing outliers, and perhaps also that the manual annotations are noisier than automated ones. For example, the ANTsBM model shows the lowest variability, with standard deviations of 61.68 mm\(^3\) (right NAc) and 63.45 mm\(^3\) (left NAc), in contrast to the values from manual annotation, which are 125.79 mm\(^3\) and 136.13 mm\(^3\).\\
Comparing the volumes of the left and right NAc, we observe that in the results of all models, the volume corresponding to the right NAc is greater than that of the left NAc, reflecting an anatomical trend preserved in both manual and automated segmentations.
One can observe a general underestimation of the NAc size, but also some patterns of bias with ANTsBM and UNesTBM displaying volumes almost 20\% smaller than the manual segmentations (Fig. \ref{fig:volumes-splitviolin}).

\begin{table*}[htp]
    \centering
    \caption{Mean and standard deviation of measured volumes for the right and left NAc (\si{\milli\meter\cubed}). Model names indicate the demographic subgroup used for training (e.g., ANTsBF was trained on the black female group). The reported statistics are calculated from applying each trained model to all the test sets.}
    \footnotesize
    \begin{tabular}{l r r r r}
        \toprule
        \textbf{Model} & \multicolumn{2}{c}{\textbf{Right NAc}} & \multicolumn{2}{c}{\textbf{Left NAc}} \\
        \cmidrule(lr){2-3} \cmidrule(lr){4-5}
        & \textbf{Mean } & \textbf{Std } & \textbf{Mean } & \textbf{Std } \\
        \midrule
        Manual & 676.97 & 125.79 & 607.13 & 136.13 \\
        \midrule
        nnU‐NetBF & 653.62 & 95.08 & 581.06 & 99.14 \\
        nnU‐NetBM & 638.20 & 115.41 & 569.97 & 108.32 \\
        nnU‐NetWF & 653.14 & 90.91 & 593.83 & 93.10 \\
        nnU‐NetWM & 665.30 & 108.89 & 604.87 & 106.29 \\
        \midrule
        CoTrBF & 658.21 & 93.08 & 582.45 & 96.79 \\
        CoTrBM & 647.76 & 119.47 & 574.53 & 114.28 \\
        CoTrWF & 664.07 & 96.01 & 600.97 & 92.19 \\
        CoTrWM & 677.96 & 111.76 & 606.37 & 109.90 \\
        \midrule
        ANTsBF & 552.27 & 68.63 & 460.58 & 67.22 \\
        ANTsBM & 491.58 & 61.68 & 437.41 & 63.45 \\
        ANTsWF & 595.83 & 70.31 & 548.00 & 66.71 \\
        ANTsWM & 618.45 & 78.61 & 577.38 & 76.94 \\
        \midrule
        UNesTBF & 614.88 & 78.02 & 528.29 & 80.83 \\
        UNesTBM & 564.65 & 86.91 & 507.03 & 85.14 \\
        UNesTWF & 623.99 & 84.35 & 558.03 & 83.48 \\
        UNesTWM & 655.31 & 94.40 & 618.32 & 87.33 \\
        \bottomrule
    \end{tabular}
    \label{table:new-volumes-combined}
\end{table*}



\subsection{Bias in volumes and segmentation performance}
In Tables~\ref{table:essp-combined-Dice} and \ref{table:essp-combined-nsd}, we observe that nnU‐Net and CoTr consistently yield the highest ESSP values in Dice coefficient and NSD, in addition to relatively balanced results across demographic groups in both metrics. However, ANTs and UNesT generally perform worse than nnU‐Net and CoTr, often by a noticeable margin. ANTs shows huge drops in ESSP for both Dice coefficient and NSD when trained on the black male or black female groups. For example, ANTs exhibits a 13\% reduction in ESSP measured by the Dice coefficient when trained on white females compared to black males. ANTs reaches its highest $\Delta$ values with 0.11 in Dice coefficient and 0.2 in NSD, respectively, when trained on black male cases in the right NAc.

We evaluated the influence of the model being same-sex, same-race, and both same-sex and race on the Dice and NSD metrics using linear mixed models. The results can be found in Tables~\ref{table:Dice-MLM-combined} and \ref{table:NSD-MLM-combined}.
The difference between same-sex and non–same-sex performance results are not statistically significant for any of the models. However, when matching race, ANTs and UNesT demonstrate significantly higher Dice coefficient than when the test and training data are not matched. When evaluated with NSD, the race-matching effect persists for ANTs, but did not reach significance for UNesT. While sex is a robust factor in volumetry (Table \ref{table:sex-race-age-effect-manual-combined}, sex matching does not significantly affect segmentation accuracy. In contrast, race matching substantially impacts segmentation.
Evaluating results between same-sex and race models (trained and tested on the same subgroup) and non-same models (race or sex mismatches), UNesT and CoTr show significantly better Dice coefficients when trained on identical race and sex sub-groups (p =0.027, 0.048 respectively). This effect is also observable for UNesT when evaluated with NSD, but not for CoTr. 
nnU‐Net was the only model that did not exhibit any changes in segmentation accuracy, considering both the Dice coefficient and NSD, across any of the three comparisons, including same race versus non-same race, same sex versus non-same sex, and the same sex and race versus non-same sex and race.

Although some models perform best on the subgroup they were trained on, there are several instances where they perform better on a different subgroup. For
example, when segmenting the right Nucleus Accumbens (NAc), the UNesT model trained on the white female (WF) dataset achieves its highest average
Dice score (0.83) on the black female (BF) test set, which is higher than its performance on the WF test set (0.81). The same case can be found in NSD results; For instance, nnU-Net trained on white male yields an NSD of 0.56 on black female subjects versus 0.54 on white male subjects. 
Detailed results for all models can be found in the supplementary material.

\begin{table*}[htp]
    \centering
    \caption{Segmentation performance metrics (DSC (overall Dice coefficient across all test sets), ESSP, $\Delta$) for right and left NAc across different models and training groups. \textbf{ESSP} (Equity-Scaled Segmentation Performance) combines overall accuracy with a penalty for cross-group disparities; \textbf{higher is better}~(\(\uparrow\)). $\Delta$ quantifies differences of each demographic group from the overall mean; \textbf{lower is better}~(\(\downarrow\)).}
    \scriptsize 
    \setlength{\tabcolsep}{3pt}   
    \renewcommand{\arraystretch}{0.95} 
    \begin{adjustbox}{max width=\textwidth}
    \begin{tabular}{l l c c c c c c c c c c c c}
        \toprule
        \multirow{2}{*}{\textbf{Structure}} & \multirow{2}{*}{\textbf{Train}} & \multicolumn{3}{c|}{\textbf{nnU-Net}} & \multicolumn{3}{c|}{\textbf{CoTr}} & \multicolumn{3}{c|}{\textbf{ANTs}} & \multicolumn{3}{c|}{\textbf{UNesT}} \\
        \cmidrule(lr){3-5} \cmidrule(lr){6-8} \cmidrule(lr){9-11} \cmidrule(lr){12-14}
         && \textbf{DSC} & ESSP~(\(\uparrow\)) & $\Delta$~(\(\downarrow\)) & \textbf{DSC} & ESSP~(\(\uparrow\)) & $\Delta$~(\(\downarrow\)) & \textbf{DSC} & ESSP~(\(\uparrow\)) & $\Delta$~(\(\downarrow\)) & \textbf{DSC} & ESSP~(\(\uparrow\)) & $\Delta$~(\(\downarrow\)) \\
        \midrule
        \multirow{4}{*}{Right NAc} & WM & 0.867 & 0.845 & 0.027 & 0.863 & 0.839 & 0.029 & 0.820 & 0.796 & 0.030 & 0.832 & 0.784 & 0.060 \\
        & WF & 0.862 & 0.838 & 0.028 & 0.859 & 0.832 & 0.032 & 0.816 & 0.793 & 0.029 & 0.817 & 0.791 & 0.032 \\
        & BM & 0.862 & 0.836 & 0.032 & 0.859 & 0.834 & 0.029 & 0.781 & 0.702 & 0.113 & 0.801 & 0.759 & 0.050 \\
        & BF & 0.862 & 0.841 & 0.025 & 0.858 & 0.836 & 0.027 & 0.792 & 0.720 & 0.100 & 0.809 & 0.780 & 0.037 \\
        \midrule
        \multirow{4}{*}{Left NAc} & WM & 0.861 & 0.849 & 0.013 & 0.856 & 0.843 & 0.016 & 0.810 & 0.794 & 0.021 & 0.825 & 0.773 & 0.066 \\
        & WF & 0.858 & 0.836 & 0.026 & 0.856 & 0.839 & 0.020 & 0.806 & 0.796 & 0.012 & 0.810 & 0.787 & 0.029 \\
        & BM & 0.854 & 0.832 & 0.026 & 0.851 & 0.831 & 0.024 & 0.758 & 0.688 & 0.102 & 0.800 & 0.748 & 0.070 \\
        & BF & 0.858 & 0.840 & 0.022 & 0.853 & 0.829 & 0.029 & 0.773 & 0.700 & 0.102 & 0.798 & 0.766 & 0.041 \\
        \bottomrule
    \end{tabular}
    \end{adjustbox}
    \label{table:essp-combined-Dice}
\end{table*}
    
\begin{table*}[htp]
    \centering
     \caption{Segmentation performance metrics (NSD (overall NSD coefficient across all test sets), ESSP, $\Delta$) for right and left NAc across different models and training groups. \textbf{ESSP} (Equity-Scaled Segmentation Performance) combines overall accuracy with a penalty for cross-group disparities; \textbf{higher is better}~(\(\uparrow\)). $\Delta$ quantifies differences of each demographic group from the overall mean; \textbf{lower is better}~(\(\downarrow\)).}
    \scriptsize 
    \setlength{\tabcolsep}{3pt}   
    \renewcommand{\arraystretch}{0.95} 
    \begin{adjustbox}{max width=\textwidth}
    \begin{tabular}{l l c c c c c c c c c c c c}
        \toprule
       \multirow{2}{*}{\textbf{Structure}} & \multirow{2}{*}{\textbf{Train}} & \multicolumn{3}{c|}{\textbf{nnU-Net}} & \multicolumn{3}{c|}{\textbf{CoTr}} & \multicolumn{3}{c|}{\textbf{ANTs}} & \multicolumn{3}{c|}{\textbf{UNesT}} \\
        \cmidrule(lr){3-5} \cmidrule(lr){6-8} \cmidrule(lr){9-11} \cmidrule(lr){12-14}
         && \textbf{NSD} & ESSP~(\(\uparrow\)) & $\Delta$~(\(\downarrow\)) & \textbf{NSD} & ESSP~(\(\uparrow\)) & $\Delta$~(\(\downarrow\)) & \textbf{NSD} & ESSP~(\(\uparrow\)) & $\Delta$~(\(\downarrow\)) & \textbf{NSD} & ESSP~(\(\uparrow\)) & $\Delta$~(\(\downarrow\)) \\
         \midrule
        \multirow{4}{*}{Right NAc} & WM & 0.527 & 0.483 & 0.090 & 0.512 & 0.469 & 0.090 & 0.430 & 0.405 & 0.060 & 0.428 & 0.387 & 0.105  \\
        & WF & 0.527 & 0.492 & 0.070 & 0.525 & 0.468 & 0.120 & 0.432 & 0.412 & 0.050 & 0.407 & 0.382 & 0.064 \\
        & BM & 0.517 & 0.457 & 0.070 & 0.510 & 0.455 & 0.120 & 0.380 & 0.316 & 0.200 & 0.392 & 0.341 & 0.1500 \\
        & BF & 0.529 & 0.500 & 0.060 & 0.525 & 0.486 & 0.080 & 0.422 & 0.364 & 0.1600 & 0.387 & 0.357 & 0.084 \\
        \midrule
        \multirow{4}{*}{Left NAc} & WM & 0.538 & 0.511 & 0.052 & 0.515 & 0.500 & 0.031 & 0.419 & 0.411 & 0.020 & 0.428 & 0.387 & 0.106 \\
        & WF & 0.522 & 0.495 & 0.055 & 0.517 & 0.495 & 0.044 & 0.424 & 0.416 & 0.020 & 0.404 & 0.371 & 0.089 \\
        & BM & 0.517 & 0.474 & 0.090 & 0.505 & 0.459 & 0.100 & 0.395 & 0.338 & 0.170 & 0.392 & 0.341 & 0.1500 \\
        & BF & 0.539 & 0.509 & 0.060 & 0.509 & 0.472 & 0.079 & 0.387 & 0.337 & 0.150 & 0.392 & 0.358 & 0.094 \\
        \bottomrule
    \end{tabular}
    \label{table:essp-combined-nsd}
    \end{adjustbox}
\end{table*}

\begin{table*}[htp]
    \centering
     \caption{Effects of Same Sex, Same Race, and Interaction on Dice coefficient for right and left NAc. $\beta_1$, $\beta_2$, and $\beta_3$. are the coefficients for a fixed factor term such as sameSex that describes the effect of the factor level on the Dice coefficient. Std Err is the standard error of the coefficient estimates.}
    \footnotesize
    \begin{tabular}{ll ccc ccc ccc}
        \toprule
        \multirow{2}{*}{\textbf{Structure}} & \multirow{2}{*}{\textbf{Model}} & \multicolumn{3}{c|}{\textbf{Same Sex}} & \multicolumn{3}{c|}{\textbf{Same Race}} & \multicolumn{3}{c|}{\textbf{Same Race $\times$ Same Sex}} \\
        \cmidrule(lr){3-5} \cmidrule(lr){6-8} \cmidrule(lr){9-11}
        & & \textbf{$\beta_2$} & \textbf{Std Err} & \textbf{P-value} & \textbf{$\beta_1$} & \textbf{Std Err} & \textbf{P-value} & \textbf{$\beta_3$} & \textbf{Std Err} & \textbf{P-value} \\
        \midrule
        \multirow{4}{*}{Right NAc} & ANTs & -0.005 & 0.006 & 0.421 & 0.021 & 0.006 & \textbf{0.000} & 0.008 & 0.008 & 0.451 \\
        & CoTr & 0.003 & 0.003 & 0.208 & 0.002 & 0.003 & 0.447 & 0.004 & 0.004 & 0.433 \\
        & nnU-Net & -0.001 & 0.003 & 0.846 & -0.000 & 0.003 & 0.979 & 0.006 & 0.004 & 0.117 \\
        & UNesT & 0.004 & 0.004 & 0.289 & 0.008 & 0.004 & \textbf{0.050} & 0.012 & 0.006 & \textbf{0.042} \\
        \midrule
        \multirow{4}{*}{Left NAc} & ANTs & -0.005 & 0.007 & 0.437 & 0.022 & 0.007 & \textbf{0.001} & 0.011 & 0.010 & 0.269 \\
        & CoTr & -0.001 & 0.003 & 0.852 & -0.000 & 0.003 & 0.986 & 0.009 & 0.004 & \textbf{0.027} \\
        & nnU-Net & 0.001 & 0.003 & 0.810 & 0.000 & 0.003 & 0.906 & 0.007 & 0.005 & 0.146 \\
        & UNesT & 0.002 & 0.005 & 0.682 & 0.011 & 0.005 & \textbf{0.030} & 0.014 & 0.007 & \textbf{0.048} \\
        \bottomrule
    \end{tabular}
    \label{table:Dice-MLM-combined}
\end{table*}

\begin{table*}[htp]
    \centering
    \caption{Effects of Same Sex, Same Race, and Interaction on NSD for right and left NAc. 
    \(\beta_2\), \(\beta_1\), and \(\beta_3\) are the coefficients for the fixed-effect terms
    Same Sex, Same Race, and their interaction (Same Race \(\times\) Same Sex), respectively. 
    “Std Err” denotes the standard error of the coefficient estimates.}
    \footnotesize
    \begin{tabular}{ll ccc ccc ccc}
        \toprule
        \multirow{2}{*}{\textbf{Structure}} & \multirow{2}{*}{\textbf{Model}} 
        & \multicolumn{3}{c|}{\textbf{Same Sex}} 
        & \multicolumn{3}{c|}{\textbf{Same Race}} 
        & \multicolumn{3}{c}{\textbf{Same Race $\times$ Same Sex}} \\
        \cmidrule(lr){3-5} \cmidrule(lr){6-8} \cmidrule(lr){9-11}
        & & \textbf{\(\beta_2\)} & \textbf{Std Err} & \textbf{P-value}
          & \textbf{\(\beta_1\)} & \textbf{Std Err} & \textbf{P-value}
          & \textbf{\(\beta_3\)} & \textbf{Std Err} & \textbf{P-value} \\
        \midrule
        \multirow{4}{*}{Right NAc} 
          & ANTs    & -0.001 & 0.010 & 0.889 & 0.040 & 0.010 & \textbf{0.000} & 0.003 & 0.015 & 0.818 \\
          & CoTr    &  0.007 & 0.007 & 0.353 & 0.009 & 0.007 & 0.248          & 0.008 & 0.010 & 0.467 \\
          & nnU-Net & -0.005 & 0.008 & 0.533 & 0.005 & 0.008 & 0.498          & 0.017 & 0.011 & 0.129 \\
          & UNesT   &  0.014 & 0.009 & 0.105 & 0.012 & 0.009 & 0.174          & 0.022 & 0.013 & 0.077 \\
        \midrule
        \multirow{4}{*}{Left NAc} 
          & ANTs    & -0.003 & 0.009 & 0.717 & 0.039 & 0.009 & \textbf{0.000} & -0.001 & 0.010 & 0.907 \\
          & CoTr    &  0.002 & 0.007 & 0.776 & 0.004 & 0.007 & 0.522          & 0.018  & 0.010 & 0.061 \\
          & nnU-Net &  0.006 & 0.008 & 0.436 & 0.009 & 0.008 & 0.259          & 0.010  & 0.011 & 0.392 \\
          & UNesT   &  0.010 & 0.010 & 0.298 & 0.017 & 0.010 & 0.070          & 0.031  & 0.013 & \textbf{0.021} \\
        \bottomrule
    \end{tabular}
    \label{table:NSD-MLM-combined}
\end{table*}

\subsection{Impact of biased segmentation on morphometric analyses}
\label{sec:morphometricanalyses}
We evaluated the influence of sex, race, and their interaction on the right NAc and left NAc volumes with a linear mixed model using manual segmentation and the different models. As shown in Table \ref{table:sex-race-effect-manual-combined}, sex and race effects can be observed in the full manual dataset (includes training and test data) on both sides, whereas in the smaller test datasets the sex effect on the right NAc volume loses significance (p=0.057). No sex-by-race interaction was observed.  
When turning to the automated biased models, one can observe a similar sex effect for all models (Table \ref{table:sex-effect-models-combined}).
The race effect, however, disappears for all models except for CoTrBF in the Left NAc (P-value=0.04). 
No automated method identified a sex-by-race interaction, which is in line with the manual segmentation results. 
Detailed tables with the race ($\gamma_1$) and sex-by-race interaction ($\gamma_3$) factors can be found in the supplementary material.
In summary, the sex effect observed in the manual segmentation remains even in the most biased models, whereas the race effect observed in the manual segmentation generally cannot be observed when segmentation is performed by highly biased models.

\begin{table*}[htbp]
    \centering
    \caption{Results for evaluating sex, race, and race $\times$ sex effects on volumes by manual annotation for right and left NAc. Coeff. is the coefficient for a fixed factor term such as Sex that describes the effect of the factor level on the volume. Std Err is the standard error of the coefficient estimates.}
    \footnotesize
    \setlength{\tabcolsep}{4pt} 
    \begin{tabular}{ll ccc ccc ccc}
        \toprule
        \multirow{2}{*}{\textbf{Structure}} & \multirow{2}{*}{\textbf{Manual}} & \multicolumn{3}{c|}{\textbf{Sex}} & \multicolumn{3}{c|}{\textbf{Race}} & \multicolumn{3}{c|}{\textbf{Race $\times$ Sex}} \\
        \cmidrule(lr){3-5} \cmidrule(lr){6-8} \cmidrule(lr){9-11}
        & & \textbf{\boldmath{$\gamma_2$}} & \textbf{Std Err} & \textbf{P-value} & \textbf{\boldmath{$\gamma_1$}} & \textbf{Std Err} & \textbf{P-value} & \textbf{\boldmath{$\gamma_3$}} & \textbf{Std Err} & \textbf{P-value} \\
        \midrule
        \multirow{2}{*}{Right NAc} &
        Manual (whole dataset) & 208.63  & 69.06  & \textbf{0.003} & 225.258 & 69.736 & \textbf{0.001} & -59.781 & 97.202  & 0.539 \\
        & Manual (Test set)      & 179.28  & 69.72  & \textbf{0.010} & 379.632 & 100.368 & \textbf{0.000} & -71.332 & 140.284 & 0.611 \\
        \midrule
        \multirow{2}{*}{Left NAc} &
        Manual (whole dataset) & 232.674 & 66.677 & \textbf{0.000} & 252.66  & 67.321 & \textbf{0.000} & 7.667   & 93.836  & 0.935 \\
        & Manual (Test set)      & 191.155 & 100.463 & 0.057 & 385.526 & 112.698 & \textbf{0.001} & -53.176 & 155.119 & 0.732 \\
        \bottomrule
    \end{tabular}
    \label{table:sex-race-effect-manual-combined}
\end{table*}

\begin{table*}[htbp]
    \centering
    \caption{Results for evaluating Sex effects on volumes by segmentation models for right and left NAc. Coeff. is the coefficient for a fixed factor term such as Sex that describes the effect of the factor level on the volume. Std Err is the standard error of the coefficient estimates, and P denotes P-value.}
    \footnotesize
    \begin{tabular}{ll ccc ccc ccc ccc}
        \toprule
        \multirow{2}{*}{\textbf{Structure}} & \multirow{2}{*}{\textbf{Model}} & \multicolumn{3}{c|}{\textbf{Trained on BF}} & \multicolumn{3}{c|}{\textbf{Trained on BM}} & \multicolumn{3}{c|}{\textbf{Trained on WF}} & \multicolumn{3}{c|}{\textbf{Trained on WM}} \\
        \cmidrule(lr){3-5} \cmidrule(lr){6-8} \cmidrule(lr){9-11} \cmidrule(lr){12-14}
        & & \textbf{\boldmath{$\gamma_2$}} & \textbf{Std Err} & \textbf{P} & \textbf{\boldmath{$\gamma_2$}} & \textbf{Std Err} & \textbf{P} & \textbf{\boldmath{$\gamma_2$}} & \textbf{Std Err} & \textbf{P} & \textbf{\boldmath{$\gamma_2$}} & \textbf{Std Err} & \textbf{P} \\
        \midrule
        \multirow{4}{*}{Right NAc} &
        ANTs & 219.8 & 49.5 & \textbf{0.000} & 171 & 41.5 & \textbf{0.000} & 131 & 50.0 & \textbf{0.009} & 214 & 58.7 & \textbf{0.000} \\
        & CoTr & 203.7 & 74.3 & \textbf{0.006} & 259.2 & 78.5 & \textbf{0.001} & 184 & 65.3 & \textbf{0.005} & 256 & 77.8 & \textbf{0.001} \\
        & nnU-Net & 231.1 & 71.5 & \textbf{0.001} & 202.4 & 74.8 & \textbf{0.007} & 166 & 74.8 & \textbf{0.026} & 248 & 78.0 & \textbf{0.001} \\
        & UNesT & 246.4 & 59.3 & \textbf{0.000} & 204 & 65.7 & \textbf{0.002} & 186 & 65.4 & \textbf{0.004} & 160 & 71.3 & \textbf{0.025} \\
        \midrule
        \multirow{4}{*}{Left NAc} &
        ANTs & 216.8 & 39.6 & \textbf{0.000} & 185 & 42.4 & \textbf{0.000} & 74.9 & 53.8 & 0.164 & 218 & 45.5 & \textbf{0.000} \\
        & CoTr & 208.8 & 82.6 & \textbf{0.012} & 164 & 83.4 & \textbf{0.049} & 168 & 69.3 & \textbf{0.015} & 142 & 77.7 & 0.066 \\
        & nnU-Net & 246.1 & 70.6 & \textbf{0.000} & 155 & 82.7 & 0.060 & 181 & 72.1 & \textbf{0.012} & 172 & 82.9 & \textbf{0.038} \\
        & UNesT & 168.6 & 65.4 & \textbf{0.010} & 145 & 65.97 & \textbf{0.027} & 158 & 61.9 & \textbf{0.010} & 101 & 73.4 & 0.166 \\
        \bottomrule
    \end{tabular}
    \label{table:sex-effect-models-combined}
\end{table*}

\subsection{Impact of Dataset Size, Demographics, and Atlas Selection on Bias in ANTs and UNesT}
In order to better understand potential sources of bias in ANTs and UNesT, we performed two additional sets of experiments. 
First, we exactly matched training dataset sizes to n=30 to rule out dataset size as a source of bias. 
Second, we established baseline settings where training data had a balanced representation of each subgroup.

For the equal sample size experiment, we mimicked the same design as above but only included 30 subjects per biased training set for training UNesT.
The race bias effect in the left NAc was statistically significant in both NSD and Dice coefficient results. 
The results can be found in Table~\ref{table:Dice-MLM-unest-only}. Notably, the $\beta$ factors are very similar to those observed in the original experiments with unequal sample sizes (Tables \ref{table:Dice-MLM-combined} and \ref{table:NSD-MLM-combined}).

For the UNesT baseline datasets, the first training set comprised 30 subjects with balanced demographics. 
Five-fold cross-validation was conducted using five folds of size six including all subgroups and two extra subjects of both of the white and black races so that the races are split evenly. 
In the second baseline experiment, UNesT was trained on 120 training subjects comprising 30 subjects from each subgroup.
We evaluated two ANTs baselines. The first one was a balanced baseline using 10 atlases: eight atlases (two per subgroup: black female, black male, white female, white male) plus two additional atlases (one black, one white) to preserve race balance. 
The second baseline with 40 atlases was composed of the exact 10 atlases from each subgroup that were used in the original biased ANTs variants. 
We compared the performance of the baseline models with the biased models.

Table~\ref{table:unest-plus-baseline-30-120-Dice} shows that the UNesT Baseline 120 model, trained on the largest and demographically diverse dataset, is the top performer, achieving the best accuracy and ESSP in the Dice coefficient and NSD. Increasing the size of a balanced training set effectively reduces bias, as evidenced by $\Delta$ for the right NAc dropping from 0.02 (30 subjects) to 0.01 (120 subjects). The same trend can also be observed in $\Delta$ by NSD dropping from 0.07 (30 subjects) to 0.03 (120 subjects). In contrast, models trained on single demographic subgroup data from black subjects consistently ranked last on all metrics. Furthermore, a clear pattern emerges where models trained on data from white subjects outperform those trained on data from black subjects, and $\Delta$ is consistently and dramatically higher for models trained on black subjects. Consequently, the ESSP is significantly lower for models trained on black subgroups in both Dice coefficient and NSD.\\
Notably, the tables~\ref{table:unest-baselines-combined-right-leftNAC-compact} and \ref{table:unest-baselines-nsd-right-left} show that the assumption that a model performs best on its matched demographic was not always true; for instance, the UNesTWF model achieved its top Dice coefficient and NSD for the right NAc when tested on the black female (BF) subgroup.

The results in table~\ref{table:ants-essp-Dice-combined}, show that ANTs variants using atlases from white subjects achieve higher accuracy and drastically lower $\Delta$ than those using atlases from black subjects in both Dice coefficient and NSD. Surprisingly, simply increasing the size of a diverse atlas set does not guarantee a fairer outcome for this traditional method. While accuracy may improve as we observe with  NSD values of Baseline 40 reaching 0.45, performance disparities can worsen, leading to a lower ESSP. This is in contrast to our findings with UNesT baselines, where larger, more diverse datasets typically mitigate bias. The segmentation performance for all the methods in Dice coefficient and NSD are shown in Tables~\ref{table:unest-baselines-combined-right-leftNAC-compact}, \ref{table:unest-baselines-nsd-right-left}, \ref{table:ants-right-left-with-permodel-and-baselines-ranked-ties} and \ref{table:ants-right-left-nsd-with-permodel-and-baselines}.

\begin{table*}[htbp]
    \centering
    \caption{Effects of Same Sex, Same Race, and Interaction on new UNesT experiment results for right and left NAc. Left block reports linear mixed-effects coefficients for DSC $\bigl(\text{DSC}= \beta_0 + \beta_1(\text{SameRace}) + \beta_2(\text{SameSex}) + \beta_3(\text{SameRace}\times\text{SameSex}) + \epsilon\bigr)$; right block reports the corresponding NSD coefficients $\bigl(\text{NSD}= \gamma_0 + \gamma_1(\text{SameRace}) + \gamma_2(\text{SameSex}) + \gamma_3(\text{SameRace}\times\text{SameSex}) + \epsilon\bigr)$. For each factor, the table lists the coefficient, its standard error (Std Err), and the P-value.}
    \label{table:Dice-MLM-unest-only}
    \setlength{\tabcolsep}{4pt}
    \renewcommand{\arraystretch}{1.1}
    \resizebox{\textwidth}{!}{%
    \begin{tabular}{l
                    ccc ccc ccc
                    c
                    ccc ccc ccc}
        \toprule
        & \multicolumn{9}{c}{\textbf{DSC (Dice) coefficients}} & \phantom{sep} & \multicolumn{9}{c}{\textbf{NSD coefficients}} \\
        \cmidrule(lr){2-10} \cmidrule(lr){12-20}
        \multirow{2}{*}{\textbf{Structure}}
        & \multicolumn{3}{c}{\textbf{Same Sex} }
        & \multicolumn{3}{c}{\textbf{Same Race} }
        & \multicolumn{3}{c}{\textbf{Same Race $\times$ Same Sex}}
        &
        & \multicolumn{3}{c}{\textbf{Same Sex} }
        & \multicolumn{3}{c}{\textbf{Same Race} }
        & \multicolumn{3}{c}{\textbf{Same Race $\times$ Same Sex} } \\
        \cmidrule(lr){2-4} \cmidrule(lr){5-7} \cmidrule(lr){8-10}
        \cmidrule(lr){12-14} \cmidrule(lr){15-17} \cmidrule(lr){18-20}
        & \makecell{\((\beta_2)\)} & Std Err & P
        & \makecell{\((\beta_1)\)} & Std Err & P
        & \makecell{\((\beta_3)\)} & Std Err & P
        &
        & \makecell{\((\gamma_2)\)} & Std Err & P
        & \makecell{\((\gamma_1)\)} & Std Err & P
        & \makecell{\((\gamma_3)\)} & Std Err & P \\
        \midrule
        Right NAc
        & 0.005 & 0.004 & 0.185
        & 0.007 & 0.004 & 0.083
        & 0.003 & 0.006 & 0.584
        &
        & 0.014 & 0.009 & 0.103
        & 0.008 & 0.009 & 0.375
        & 0.008 & 0.012 & 0.526 \\
        Left NAc
        & 0.002 & 0.004 & 0.587
        & 0.014 & 0.004 & \textbf{0.001}
        & 0.004 & 0.006 & 0.546
        &
        & 0.009 & 0.008 & 0.231
        & 0.022 & 0.008 & \textbf{0.005}
        & 0.007 & 0.011 & 0.542 \\
        \bottomrule
    \end{tabular}
    }
\end{table*}

\begin{table*}[htp]
\centering
\footnotesize
\caption{Segmentation performance metrics and fairness metric based on Dice coefficient
for biased UNesT models for right and left NAc. (1 shows best) (\textbf{ESSP} (Equity-Scaled Segmentation Performance) combines overall
accuracy with a penalty for cross-group disparities; \textbf{higher is better}~(\(\uparrow\)).
\(\boldsymbol{\Delta}\) quantifies differences of each demographic group from the overall mean; \textbf{lower is better}~(\(\downarrow\)).)}
\label{table:unest-plus-baseline-30-120-Dice}
\setlength{\tabcolsep}{3pt}    
\renewcommand{\arraystretch}{0.92} 
\begin{tabular}{l l c c c c c c}
\toprule
\textbf{Structure} & \textbf{Train} & \textbf{DSC} & ESSP(\(\uparrow\)) &  $\Delta$(\(\downarrow\)) & \textbf{NSD} & ESSP(\(\uparrow\)) & $\Delta$(\(\downarrow\)) \\
\midrule
\multirow{6}{*}{Right NAc}
  & WM    & 0.81\textsuperscript{2} & 0.79\textsuperscript{2} & 0.02\textsuperscript{2}  &
  0.41\textsuperscript{2} & 0.39\textsuperscript{2} & 0.04\textsuperscript{2} \\
  & WF    & 0.81\textsuperscript{2} & 0.78\textsuperscript{3} & 0.03\textsuperscript{3} &
  0.41\textsuperscript{2} & 0.37\textsuperscript{4} & 0.08\textsuperscript{4}\\
  & BM    & 0.80\textsuperscript{3} & 0.76\textsuperscript{4} & 0.04\textsuperscript{4} &
  0.37\textsuperscript{5} & 0.34\textsuperscript{6} & 0.09\textsuperscript{5}\\
  & BF    & 0.80\textsuperscript{3} & 0.78\textsuperscript{3} & 0.03\textsuperscript{3} &
  0.39\textsuperscript{4} & 0.36\textsuperscript{5} & 0.07\textsuperscript{3}\\
  & Baseline30  & 0.81\textsuperscript{2} & 0.79\textsuperscript{2} & 0.02\textsuperscript{2} & 
  0.40\textsuperscript{3} & 0.38\textsuperscript{3} & 0.07\textsuperscript{3}\\
  & Baseline120 & \textbf{0.82}\textsuperscript{1} & \textbf{0.81}\textsuperscript{1} & \textbf{0.01}\textsuperscript{1} & \textbf{0.42}\textsuperscript{1} & \textbf{0.40}\textsuperscript{1} & \textbf{0.03}\textsuperscript{1} \\
\midrule
\multirow{6}{*}{Left NAc}
  & WM    & \textbf{0.81}\textsuperscript{1} & \textbf{0.80}\textsuperscript{1} & \textbf{0.00}\textsuperscript{1} & 0.40\textsuperscript{2} & 0.39\textsuperscript{2} & \textbf{0.02}\textsuperscript{1} \\
  & WF    & 0.80\textsuperscript{2} & 0.79\textsuperscript{2} & 0.02\textsuperscript{3} & 0.40\textsuperscript{2} & 0.38\textsuperscript{3} & 0.06\textsuperscript{3}\\
  & BM    & 0.79\textsuperscript{3} & 0.74\textsuperscript{4} & 0.06\textsuperscript{5} &
  0.39\textsuperscript{3} & 0.34\textsuperscript{5} & 0.15\textsuperscript{5}\\
  & BF    & 0.79\textsuperscript{3} & 0.75\textsuperscript{3} & 0.05\textsuperscript{4} &
  0.38\textsuperscript{4} & 0.34\textsuperscript{5} & 0.11\textsuperscript{4}\\
  & Baseline30  & 0.80\textsuperscript{2} & 0.79\textsuperscript{2} & 0.01\textsuperscript{2} & 0.40\textsuperscript{2} & 0.38\textsuperscript{4} & 0.06\textsuperscript{3} \\
  & Baseline120 & \textbf{0.81}\textsuperscript{1} & \textbf{0.80}\textsuperscript{1} & 0.01\textsuperscript{2} &\textbf{0.42}\textsuperscript{1} & \textbf{0.40}\textsuperscript{1} & 0.04\textsuperscript{2} \\
\bottomrule
\end{tabular}
\end{table*}

\begin{table*}[htp]
\centering
\caption{Average and standard deviation of Dice coefficient (Avg$\pm$Std) for right and left NAc. Columns: UNesT trained on each subgroup (WM/WF/BM/BF -- matched sample sizes) and Baselines. Superscripts rank average within each row ( 1 shows best).}
\label{table:unest-baselines-combined-right-leftNAC-compact}
\setlength{\tabcolsep}{3.5pt}
\renewcommand{\arraystretch}{0.98}
\footnotesize
\begin{tabular}{llcccccc}
\toprule
\textbf{Structure} & \textbf{Test} & \textbf{UNesTWM} & \textbf{UNesTWF} & \textbf{UNesTBM} & \textbf{UNesTBF} & \textbf{Baseline (30)} & \textbf{Baseline (120)} \\
\midrule
\multirow{4}{*}{Right NAc}
& WM & \textbf{0.83}\textsuperscript{1}\,$\pm$\,0.02 & 0.81\textsuperscript{3}\,$\pm$\,0.02 & 0.80\textsuperscript{4}\,$\pm$\,0.03 & 0.80\textsuperscript{4}\,$\pm$\,0.04 & 0.81\textsuperscript{3}\,$\pm$\,0.04 & 0.82\textsuperscript{2}\,$\pm$\,0.03 \\
& WF & 0.81\textsuperscript{2}\,$\pm$\,0.04 & 0.81\textsuperscript{2}\,$\pm$\,0.03 & 0.78\textsuperscript{4}\,$\pm$\,0.04 & 0.79\textsuperscript{3}\,$\pm$\,0.04 & 0.81\textsuperscript{2}\,$\pm$\,0.04 & \textbf{0.82}\textsuperscript{1}\,$\pm$\,0.03 \\
& BM & 0.80\textsuperscript{3}\,$\pm$\,0.05 & 0.81\textsuperscript{2}\,$\pm$\,0.04 & 0.81\textsuperscript{2}\,$\pm$\,0.03 & 0.80\textsuperscript{3}\,$\pm$\,0.04 & \textbf{0.82}\textsuperscript{1}\,$\pm$\,0.03 & 0.81\textsuperscript{2}\,$\pm$\,0.04 \\
& BF & 0.81\textsuperscript{3}\,$\pm$\,0.04 & \textbf{0.83}\textsuperscript{1}\,$\pm$\,0.03 & 0.80\textsuperscript{4}\,$\pm$\,0.03 & 0.82\textsuperscript{2}\,$\pm$\,0.03 & 0.82\textsuperscript{2}\,$\pm$\,0.03 & \textbf{0.83}\textsuperscript{1}\,$\pm$\,0.03 \\
\midrule
\multirow{4}{*}{Left NAc}
& WM & \textbf{0.82}\textsuperscript{1}\,$\pm$\,0.02 & 0.80\textsuperscript{3}\,$\pm$\,0.04 & 0.79\textsuperscript{4}\,$\pm$\,0.04 & 0.79\textsuperscript{4}\,$\pm$\,0.03 & 0.80\textsuperscript{3}\,$\pm$\,0.04 & 0.81\textsuperscript{2}\,$\pm$\,0.03 \\
& WF & 0.81\textsuperscript{1}\,$\pm$\,0.03 & 0.80\textsuperscript{2}\,$\pm$\,0.02 & 0.77\textsuperscript{5}\,$\pm$\,0.03 & 0.78\textsuperscript{4}\,$\pm$\,0.03 & 0.79\textsuperscript{3}\,$\pm$\,0.03 & \textbf{0.81}\textsuperscript{1}\,$\pm$\,0.02 \\
& BM & 0.80\textsuperscript{2}\,$\pm$\,0.05 & 0.80\textsuperscript{2}\,$\pm$\,0.05 & \textbf{0.81}\textsuperscript{1}\,$\pm$\,0.03 & 0.80\textsuperscript{2}\,$\pm$\,0.04 & 0.80\textsuperscript{2}\,$\pm$\,0.05 & \textbf{0.81}\textsuperscript{1}\,$\pm$\,0.05 \\
& BF & 0.81\textsuperscript{2}\,$\pm$\,0.03 & 0.81\textsuperscript{2}\,$\pm$\,0.03 & 0.81\textsuperscript{2}\,$\pm$\,0.03 & \textbf{0.82}\textsuperscript{1}\,$\pm$\,0.03 & 0.80\textsuperscript{3}\,$\pm$\,0.05 & \textbf{0.82}\textsuperscript{1}\,$\pm$\,0.03 \\
\bottomrule
\end{tabular}
\end{table*}

\begin{table*}[htp]
\centering
\caption{Average and standard deviation of NSD (Avg$\pm$Std) for right and left NAc. Columns: UNesT trained on each subgroup (WM/WF/BM/BF -- matched sample sizes) and Baselines. Superscripts rank average within each row ( 1 shows best).}
\label{table:unest-baselines-nsd-right-left}
\footnotesize
\setlength{\tabcolsep}{3.5pt}
\renewcommand{\arraystretch}{0.95}
\begin{tabular}{llcccccc}
\toprule
\textbf{Structure} & \textbf{Test} & \textbf{UNesTWM} & \textbf{UNesTWF} & \textbf{UNesTBM} & \textbf{UNesTBF} & \textbf{Baseline (30)} & \textbf{Baseline (120)} \\
\midrule
\multirow{4}{*}{Right NAc}
& WM & \textbf{0.42}\textsuperscript{1}\,$\pm$\,0.05 & 0.39\textsuperscript{2}\,$\pm$\,0.05 & 0.37\textsuperscript{4}\,$\pm$\,0.06 & 0.38\textsuperscript{3}\,$\pm$\,0.08 & 0.38\textsuperscript{3}\,$\pm$\,0.07 & \textbf{0.42}\textsuperscript{1}\,$\pm$\,0.05 \\
& WF & 0.40\textsuperscript{2}\,$\pm$\,0.08 & 0.38\textsuperscript{4}\,$\pm$\,0.05 & 0.33\textsuperscript{6}\,$\pm$\,0.05 & 0.36\textsuperscript{5}\,$\pm$\,0.07 & 0.39\textsuperscript{3}\,$\pm$\,0.07 & \textbf{0.41}\textsuperscript{1}\,$\pm$\,0.07 \\
& BM & 0.39\textsuperscript{4}\,$\pm$\,0.08 & 0.40\textsuperscript{3}\,$\pm$\,0.08 & 0.41\textsuperscript{2}\,$\pm$\,0.06 & 0.38\textsuperscript{5}\,$\pm$\,0.08 & \textbf{0.42}\textsuperscript{1}\,$\pm$\,0.06 & 0.41\textsuperscript{2}\,$\pm$\,0.08 \\
& BF & 0.42\textsuperscript{4}\,$\pm$\,0.07 & \textbf{0.45}\textsuperscript{1}\,$\pm$\,0.07 & 0.38\textsuperscript{5}\,$\pm$\,0.08 & 0.43\textsuperscript{3}\,$\pm$\,0.07 & 0.42\textsuperscript{4}\,$\pm$\,0.08 & 0.44\textsuperscript{2}\,$\pm$\,0.07 \\
\midrule
\multirow{4}{*}{Left NAc}
& WM & 0.39\textsuperscript{2}\,$\pm$\,0.06 & 0.39\textsuperscript{2}\,$\pm$\,0.08 & 0.35\textsuperscript{3}\,$\pm$\,0.08 & 0.35\textsuperscript{3}\,$\pm$\,0.07 & 0.39\textsuperscript{2}\,$\pm$\,0.06 & \textbf{0.40}\textsuperscript{1}\,$\pm$\,0.07 \\
& WF & 0.40\textsuperscript{2}\,$\pm$\,0.05 & 0.40\textsuperscript{2}\,$\pm$\,0.06 & 0.35\textsuperscript{4}\,$\pm$\,0.04 & 0.35\textsuperscript{4}\,$\pm$\,0.05 & 0.38\textsuperscript{3}\,$\pm$\,0.06 & \textbf{0.42}\textsuperscript{1}\,$\pm$\,0.04 \\
& BM & 0.40\textsuperscript{4}\,$\pm$\,0.09 & 0.39\textsuperscript{5}\,$\pm$\,0.09 & \textbf{0.43}\textsuperscript{1}\,$\pm$\,0.08 & 0.38\textsuperscript{6}\,$\pm$\,0.07 & 0.41\textsuperscript{3}\,$\pm$\,0.08 & 0.42\textsuperscript{2}\,$\pm$\,0.08 \\
& BF & 0.42\textsuperscript{4}\,$\pm$\,0.06 & 0.44\textsuperscript{2}\,$\pm$\,0.06 & 0.43\textsuperscript{3}\,$\pm$\,0.07 & 0.44\textsuperscript{2}\,$\pm$\,0.08 & 0.41\textsuperscript{5}\,$\pm$\,0.09 & \textbf{0.45}\textsuperscript{1}\,$\pm$\,0.07 \\
\bottomrule
\end{tabular}
\end{table*}

\begin{table*}[htp]
    \centering
    \footnotesize
    \caption{Fairness metrics based on Dice coefficient for ANTs and baselines (1 shows the best)(\textbf{ESSP} (Equity-Scaled Segmentation Performance) combines overall accuracy with a penalty for cross-group disparities; \textbf{higher is better}~(\(\uparrow\)).
\(\boldsymbol{\Delta}\) quantifies differences of each demographic group from the overall mean; \textbf{lower is better}~(\(\downarrow\)).)}
    \setlength{\tabcolsep}{3pt}    
    \renewcommand{\arraystretch}{0.92} 
    \begin{tabular}{l l c c c c c c}
        \toprule
        \textbf{Structure} & \textbf{Train} & \textbf{DSC} & ESSP(\(\uparrow\)) &  $\Delta$(\(\downarrow\)) & \textbf{NSD} & ESSP(\(\uparrow\)) & $\Delta$(\(\downarrow\)) \\
        \midrule

        \multirow{6}{*}{Right NAc}
        & WM & \textbf{0.82}\textsuperscript{1} & \textbf{0.79}\textsuperscript{1} & 0.03\textsuperscript{2} &
        0.43\textsuperscript{2} & 0.40\textsuperscript{2} & 0.06\textsuperscript{2}\\
        & WF & 0.81\textsuperscript{2} & \textbf{0.79}\textsuperscript{1} & \textbf{0.02}\textsuperscript{1} & 0.43\textsuperscript{2} & \textbf{0.41}\textsuperscript{1} & \textbf{0.05}\textsuperscript{1} \\
        & BM & 0.78\textsuperscript{4} & 0.70\textsuperscript{3} & 0.11\textsuperscript{5} & 0.38\textsuperscript{6} & 0.31\textsuperscript{6} & 0.20\textsuperscript{6} \\
        & BF & 0.79\textsuperscript{3} & 0.72\textsuperscript{2} & 0.10\textsuperscript{4} &  0.42\textsuperscript{4} & 0.36\textsuperscript{4} & 0.16\textsuperscript{5} \\
        & Baseline10  & 0.78\textsuperscript{4} & 0.72\textsuperscript{2} & 0.08\textsuperscript{3} &  0.40\textsuperscript{5} & 0.35\textsuperscript{5} & 0.12\textsuperscript{3} \\
        & Baseline40  & 0.81\textsuperscript{2} & 0.72\textsuperscript{2} & 0.12\textsuperscript{6} &
        \textbf{0.45}\textsuperscript{1} & 0.40\textsuperscript{2} & 0.13\textsuperscript{4}\\
        \midrule

        \multirow{6}{*}{Left NAc}
        & WM & \textbf{0.81}\textsuperscript{1} & \textbf{0.79}\textsuperscript{1} & 0.02\textsuperscript{2} &
        0.41\textsuperscript{3} & \textbf{0.41}\textsuperscript{1} & \textbf{0.02}\textsuperscript{1}\\    
        & WF & 0.80\textsuperscript{2} & \textbf{0.79}\textsuperscript{1} & \textbf{0.01}\textsuperscript{1} &
        0.42\textsuperscript{2} & 0.41\textsuperscript{2} & \textbf{0.02}\textsuperscript{1}\\
        & BM & 0.75\textsuperscript{5} & 0.68\textsuperscript{4} & 0.10\textsuperscript{4} & 
        0.39\textsuperscript{4} & 0.33\textsuperscript{5} & 0.17\textsuperscript{6}\\
        & BF & 0.77\textsuperscript{4} & 0.70\textsuperscript{3} & 0.10\textsuperscript{4} &
        0.38\textsuperscript{6} & 0.33\textsuperscript{5} & 0.15\textsuperscript{5}\\
        & Baseline 10  & 0.78\textsuperscript{3} & 0.72\textsuperscript{2} & 0.07\textsuperscript{3} &
        0.39\textsuperscript{4} & 0.35\textsuperscript{4} & 0.14\textsuperscript{4}\\
        & Baseline 40  & 0.80\textsuperscript{2} & 0.72\textsuperscript{2} & 0.11\textsuperscript{6} &
        \textbf{0.45}\textsuperscript{1} & 0.39\textsuperscript{3} & 0.13\textsuperscript{3} \\
        \bottomrule
    \end{tabular}
    \label{table:ants-essp-Dice-combined}
\end{table*}

\begin{table*}[htp]
\centering
\caption{Average and standard deviation of Dice coefficient (Avg$\pm$Std) for right and left NAc. Columns: ANTs trained on each subgroup (WM/WF/BM/BF) and Baselines. Superscripts rank average within each row ( 1 shows best).}
\label{table:ants-right-left-with-permodel-and-baselines-ranked-ties}
\footnotesize
\setlength{\tabcolsep}{3.5pt}
\renewcommand{\arraystretch}{0.95}
\begin{tabular}{llcccccc}
\toprule
\textbf{Structure} & \textbf{Test} & \textbf{ANTsWM} & \textbf{ANTsWF} & \textbf{ANTsBM} & \textbf{ANTsBF} & \textbf{Baseline (10)} & \textbf{Baseline (40)} \\
\midrule
\multirow{4}{*}{Right NAc}
& WM & \textbf{0.82}\textsuperscript{1}$\pm$0.02 & 0.81\textsuperscript{3}$\pm$0.03 & 0.76\textsuperscript{6}$\pm$0.05 & 0.77\textsuperscript{5}$\pm$0.06 & 0.78\textsuperscript{4}$\pm$0.04 & 0.81\textsuperscript{2}$\pm$0.04 \\
& WF & \textbf{0.81}\textsuperscript{1}$\pm$0.04 & 0.80\textsuperscript{2}$\pm$0.04 & 0.74\textsuperscript{6}$\pm$0.04 & 0.75\textsuperscript{3}$\pm$0.05 & 0.74\textsuperscript{5}$\pm$0.05 & 0.74\textsuperscript{4}$\pm$0.18 \\
& BM & 0.80\textsuperscript{5}$\pm$0.04 & 0.81\textsuperscript{2}$\pm$0.04 & 0.81\textsuperscript{2}$\pm$0.03 & 0.81\textsuperscript{2}$\pm$0.05 & 0.79\textsuperscript{6}$\pm$0.03 & \textbf{0.83}\textsuperscript{1}$\pm$0.03 \\
& BF & 0.82\textsuperscript{2}$\pm$0.04 & 0.82\textsuperscript{2}$\pm$0.04 & 0.80\textsuperscript{6}$\pm$0.03 & 0.82\textsuperscript{2}$\pm$0.04 & 0.81\textsuperscript{5}$\pm$0.04 & \textbf{0.84}\textsuperscript{1}$\pm$0.03 \\
\midrule
\multirow{4}{*}{Left NAc}
& WM & \textbf{0.82}\textsuperscript{1}$\pm$0.03 & 0.81\textsuperscript{2}$\pm$0.03 & 0.73\textsuperscript{6}$\pm$0.06 & 0.76\textsuperscript{5}$\pm$0.06 & 0.77\textsuperscript{4}$\pm$0.05 & 0.80\textsuperscript{3}$\pm$0.05 \\
& WF & \textbf{0.80}\textsuperscript{1}$\pm$0.04 & 0.80\textsuperscript{1}$\pm$0.03 & 0.72\textsuperscript{5}$\pm$0.05 & 0.72\textsuperscript{5}$\pm$0.07 & 0.74\textsuperscript{3}$\pm$0.05 & 0.74\textsuperscript{4}$\pm$0.18 \\
& BM & 0.80\textsuperscript{3}$\pm$0.06 & 0.80\textsuperscript{3}$\pm$0.06 & 0.78\textsuperscript{6}$\pm$0.06 & 0.79\textsuperscript{5}$\pm$0.06 & 0.80\textsuperscript{2}$\pm$0.06 & \textbf{0.82}\textsuperscript{1}$\pm$0.05 \\
& BF & 0.80\textsuperscript{3}$\pm$0.06 & 0.80\textsuperscript{3}$\pm$0.06 & 0.77\textsuperscript{6}$\pm$0.03 & 0.80\textsuperscript{3}$\pm$0.03 & 0.80\textsuperscript{2}$\pm$0.03 & \textbf{0.83}\textsuperscript{1}$\pm$0.03 \\
\bottomrule
\end{tabular}
\end{table*}

\begin{table*}[htp]
\centering
\footnotesize
\caption{Average and standard deviation of NSD (Avg$\pm$Std) for right and left NAc. Columns: ANTs trained on each subgroup (WM/WF/BM/BF) and Baselines. Superscripts rank average within each row ( 1 shows best). }
\label{table:ants-right-left-nsd-with-permodel-and-baselines}
\scriptsize
\setlength{\tabcolsep}{3.5pt}
\renewcommand{\arraystretch}{0.95}
\begin{tabular}{llcccccc}
\toprule
\textbf{Structure} & \textbf{Test} & \textbf{ANTsWM} & \textbf{ANTsWF} & \textbf{ANTsBM} & \textbf{ANTsBF} & \textbf{Baseline (10)} & \textbf{Baseline (40)} \\
\midrule
\multirow{4}{*}{Right NAc}
& WM & 0.44\textsuperscript{2}$\pm$0.05 & 0.43\textsuperscript{3}$\pm$0.08 & 0.35\textsuperscript{6}$\pm$0.07 & 0.38\textsuperscript{5}$\pm$0.12 & 0.38\textsuperscript{4}$\pm$0.08 & \textbf{0.44}\textsuperscript{1}$\pm$0.08 \\
& WF & \textbf{0.43}\textsuperscript{1}$\pm$0.07 & 0.41\textsuperscript{2}$\pm$0.07 & 0.31\textsuperscript{6}$\pm$0.06 & 0.36\textsuperscript{4}$\pm$0.06 & 0.35\textsuperscript{5}$\pm$0.06 & 0.40\textsuperscript{3}$\pm$0.06 \\
& BM & 0.40\textsuperscript{6}$\pm$0.05 & 0.44\textsuperscript{2}$\pm$0.07 & 0.43\textsuperscript{4}$\pm$0.07 & 0.44\textsuperscript{2}$\pm$0.10 & 0.41\textsuperscript{5}$\pm$0.07 & \textbf{0.47}\textsuperscript{1}$\pm$0.08 \\
& BF & 0.45\textsuperscript{3}$\pm$0.08 & 0.45\textsuperscript{3}$\pm$0.08 & 0.43\textsuperscript{6}$\pm$0.08 & 0.46\textsuperscript{2}$\pm$0.09 & 0.44\textsuperscript{5}$\pm$0.08 & \textbf{0.50}\textsuperscript{1}$\pm$0.09 \\
\midrule
\multirow{4}{*}{Left NAc}
& WM & \textbf{0.42}\textsuperscript{1}$\pm$0.07 & \textbf{0.42}\textsuperscript{1}$\pm$0.07 & 0.35\textsuperscript{6}$\pm$0.07 & 0.36\textsuperscript{5}$\pm$0.10 & 0.37\textsuperscript{4}$\pm$0.09 & 0.41\textsuperscript{3}$\pm$0.10 \\
& WF & 0.42\textsuperscript{2}$\pm$0.07 & \textbf{0.43}\textsuperscript{1}$\pm$0.06 & 0.33\textsuperscript{6}$\pm$0.06 & 0.34\textsuperscript{5}$\pm$0.08 & 0.35\textsuperscript{4}$\pm$0.06 & 0.41\textsuperscript{3}$\pm$0.07 \\
& BM & 0.41\textsuperscript{6}$\pm$0.09 & 0.42\textsuperscript{5}$\pm$0.08 & 0.43\textsuperscript{3}$\pm$0.07 & 0.43\textsuperscript{3}$\pm$0.10 & 0.43\textsuperscript{2}$\pm$0.10 & \textbf{0.48}\textsuperscript{1}$\pm$0.10 \\
& BF & 0.43\textsuperscript{3}$\pm$0.09 & 0.43\textsuperscript{3}$\pm$0.09 & 0.42\textsuperscript{5}$\pm$0.06 & 0.42\textsuperscript{5}$\pm$0.07 & 0.43\textsuperscript{2}$\pm$0.07 & \textbf{0.48}\textsuperscript{1}$\pm$0.07 \\
\bottomrule
\end{tabular}
\end{table*}

\section{Discussion}
Our investigation highlights how an imbalance in demographic factors such as race and sex influences the segmentation quality and the volumetric measurements of the NAcs. 
While all models preserved the anatomical trend of larger right NAc volumes compared to the left NAc, aligning with manual segmentations, most models exhibited narrower volume standard deviations than manual annotations. 
An important finding in this study is that while most models faithfully preserve the sex-based volume differences seen in the manually labeled ground-truth data, race-based differences present in the manually annotated data vanish in all automated biased models. 

From a fairness perspective, ESSPs measured by both Dice coefficient and NSD indicate that nnU‐Net and CoTr often achieve the highest accuracy, combined with smaller inter‐group disparities. In contrast, ANTs is highly sensitive to the race of the training set, with significantly lower Dice coefficient and NSD and larger $\Delta$ values measured by both Dice coefficient and NSD for models trained on black subgroups compared to white subgroups. While UNesT outperforms ANTs in terms of delivering higher overall Dice coefficient and NSD and more consistent performance across demographic groups, it has inferior performance compared to nnU‐Net and CoTr in achieving consistent accuracy across all demographic groups.
The linear mixed model results further show that among the factors influencing segmentation accuracy race-matching between training and test datasets provides a substantial performance benefit for the ANTs and UNesT models. Perhaps surprisingly, sex-matching had far less effect on performance.
nnU-Net is the only model whose performance was not influenced by race-matching or sex and race-matching. Our results of a strong race bias effect align with the insights in the study by \citet{ioannou2022studydemographicbiascnnbased}, which reported that the race bias effect was more significant than the sex bias effect.
We also compared the performance of methods using NSD and found that the overall ranking of the models remained similar to the Dice results in terms of fairness, with nnU-Net as the top performer and ANTs and UNesT showing the most vulnerability to bias. The magnitude of bias was amplified with NSD. For example, $\Delta$ values for ANTs soared to 0.20. For the CoTr model, while it appeared both accurate and fair, its performance disparities were much more pronounced when evaluated with NSD. NSD amplifies boundary-level inequities that the Dice coefficient can conceal.
Our analysis also revealed some counterintuitive patterns where some models trained on one demographic subgroup occasionally perform better on another. For instance, the UNesT model trained on white females (WF) achieved a higher average Dice coefficient on the black female (BF) test set than on its own WF test set.

For the UNesT model, we conducted additional experiments with balanced-size training sets, and the race bias effect was observed in both NSD and Dice coefficient results. Further baseline experiments with demographically balanced datasets suggest that race-balanced, larger datasets significantly mitigate unfairness. 
The results of similar ANTs baseline experiments with balanced atlases highlight the complex nature of bias and suggest that simply balancing demographic attributes in the atlases does not equate to improving fairness. This is in contrast to our findings with UNesT baselines, where larger, more diverse datasets typically mitigate bias.

These findings can have important implications. Biased segmentation models can misrepresent brain structures. For example, ANTs trained on black males shows substantial under-segmentation of the left NAc, with volumes nearly 28\% smaller than manual annotations. This difference can influence clinical applications as right and left NAc can serve as a biomarker. For example, Major depressive disorder has been associated with persistent reductions in NAc volume~\citep{Ancelin2019}.

It is worth noting that while the existence of bias in deep learning models is well-established, its manifestation within brain segmentation can be subtle and highly variable across different methods. This variability necessitates quantitative assessment of fairness in neuroimaging studies. Our findings corroborate this, demonstrating that bias can be pronounced in certain architectures like UNesT, yet not observable in frameworks such as nnU-Net.
In brain MRI segmentation, prior bias studies typically assess a single deep-learning model, using low-quality labels as ground truth for training datasets~\citep{ioannou2022studydemographicbiascnnbased}. Outside the brain, multi-model comparisons have been reported~\citep{Lee2023Cardiac}. Our study is, to our knowledge, the first in brain MRI to compare deep learning and a classical atlas-based method with respect to bias.

One of the challenges of our approach is that it is difficult to pinpoint the source of the observed biases. 
Indeed, we chose to evaluate how models perform, as recommended ``off-the-shelf" by their creators, as our goal was to investigate bias in methods that are not just standalone architectures, but complete pipelines, each with hundreds of parameters and author-recommended configurations that are integral to their performance. This is in contrast to using a very restricted framework in which we changed only a few parameters and made an inference about the source of these biases. 
While pinpointing the source of bias in segmentation models is complex, we can compare the methods' inherent characteristics. The ANTs method is vulnerable to bias because its weighted-voting strategy produces systematic errors when the atlas set can be dominated by a single demographic's anatomy. Conversely, nnU-Net's superior fairness likely stems from its adaptive data augmentation, which forces the model to learn generalizable anatomical patterns instead of demographic-related features. We hypothesize that UNesT, despite its powerful transformer architecture, was more susceptible to overfitting on the demographic traits of its small training set due to a lack of such rigorous data augmentation.

A limitation of this work is its relatively small dataset size within each demographic subgroup, which may restrict generalizability and more nuanced biases. Moreover, while our study focused primarily on adult populations, biases may appear differently in children or older people. Additionally, our study focused solely on the right and left nucleus accumbens which are small structures. Future research should examine a broader range of structures and also other datasets to confirm whether this trend persists. 
We also acknowledge that our study is restricted to a single structure. Further study will need to extend bias evaluation beyond the NAc. The main challenge here is the lack of benchmark datasets with ground truth segmentations across the brain.
Although our study focused solely on the effects of demographic attributes such as sex and race, we recognize that these are not the only potential sources of bias within a dataset. Other factors can also have a considerable impact on brain volumes and model segmentation performance. For instance, the HCP Young Adult dataset includes participants aged 22 to 35 and therefore excludes other age groups, such as children and older adults. As noted in our limitations, biases may manifest differently in children or older adults. In addition, the HCP dataset comprises only healthy subjects, which means that the findings may not generalize to individuals with psychiatric disorders, congenital abnormalities in brain anatomy, or other clinical conditions. Furthermore, social status and education level can also be confounding factors that are not considered in our study. Technical biases stemming from variations in scanner hardware, software, or acquisition parameters also introduce systematic variations. 
Additionally, although we intentionally designed our study to isolate demographic effects, this approach does not reflect real-world scenarios where training sets are more heterogeneous, potentially amplifying the observed biases.

Finally, our study is diagnostic in nature and does not test potential bias mitigation strategies suitable for unbalanced data regimes. For instance, sensitive class-aware data augmentation~\citep{Xu2024} could be employed. This technique involves applying more aggressive data augmentation to underrepresented demographic groups within the training set, thereby encouraging the model to learn more robust and generalizable features that are not dependent on sensitive attributes. To address subgroup imbalances, data synthesis can be utilized. This approach leverages generative models to augment the training dataset with synthetic data for the minority class, ensuring a more balanced distribution for model training~\citep{POMBO2023102723}. Incorporating such prescriptive strategies in future work is a critical next step toward developing segmentation tools that are not only accurate but also fair across diverse populations.
Finally, evaluating changes in bias under different network architectures or training procedures could inform best practices for equitable brain MRI segmentation.

\section{Conclusion}
This study provides insights into demographic biases in brain MRI segmentation. 
Our results show a nuanced picture with different methods displaying different levels of sensitivity to demographic biases in their training data. 
ANTs and UNesT were most affected while nnU-Net seemed to be the most robust to biases.
In terms of the relative importance of demographic variables, race seemed to impact segmentation performance more than sex, and most models show a lower overall segmentation accuracy and ESSP in both Dice coefficient and NSD when trained on datasets from black demographic groups than those trained on white demographic groups.
Additionally, we found that these performance biases impact morphometric studies. Notably, a race effect on NAc volumes was observed with manual segmentations, but was not observed with automated methods trained with biased models.
These findings underscore the need for diverse training sets and rigorous model assessments across multiple demographic strata to achieve truly equitable and clinically reliable brain MRI segmentation.
Finally, our study remains limited in scope as our results are based on studying one anatomical structure and a single dataset. 
Further research is required to conduct a more comprehensive investigation to determine whether these results are generalizable across diverse structures and datasets.

\acks{Data were provided by the Human Connectome Project, WU-Minn Consortium (Principal Investigators: David
Van Essen and Kamil Ugurbil; 1U54MH091657) funded by the 16 NIH Institutes and Centers that support the
NIH Blueprint for Neuroscience Research; and by the McDonnell Center for Systems Neuroscience at Washington
University. This work was supported by Natural Sciences and Engineering Research Council of Canada grants Discovery grant RGPIN-2023-05443 and Canada Research Chair CRC-2022-00183.}

%
\ethics{The work follows appropriate ethical standards in conducting research and writing the manuscript, following all applicable laws and regulations regarding treatment of animals or human subjects.}

\coi{We declare we don't have conflicts of interest.}

\data{
The data supporting the findings of this study are publicly available. 
The full HCP Young Adult dataset is accessible at:
{\footnotesize \url{https://www.humanconnectome.org/study/hcp-young-adult}}

The manual segmentations are available on Zenodo: \footnotesize{\url{https://doi.org/10.5281/zenodo.17336059}}
}

\bibliography{melba-sample}





\end{document}